\RequirePackage{etoolbox}
\csdef{input@path}{%
 {sty/}
 {img/}
}%
\csgdef{bibdir}{bib/}

\documentclass[ba]{imsart}
\pubyear{0000}
\volume{00}
\issue{0}
\doi{0000}
\firstpage{1}
\lastpage{1}

\usepackage{amsthm}
\usepackage{amsmath}
\usepackage{natbib}
\usepackage{changepage}
\usepackage[colorlinks,citecolor=blue,urlcolor=blue,filecolor=blue,backref=page]{hyperref}
\usepackage{graphicx}
\usepackage{tablefootnote}
\usepackage{threeparttable}
\startlocaldefs
\numberwithin{equation}{section}
\theoremstyle{plain}

\endlocaldefs

\begin{document}

\begin{frontmatter}
\title{Approximation Properties of Variational Bayes for Vector Autoregressions \thanksref{T1}}
\thankstext{T1}{I Would like to thank Bill Griffiths and Tomasz Wo\'zniak for their very useful comments on the paper}

\begin{aug}
\author{\fnms{Reza Hajargasht} \snm{}\thanksref{addr1,t1,t2,m1}\ead[label=e1]{rhajargasht@swin.edu.au}}

\runtitle{VB Approximation Error}

\address[addr1]{Author's Address: Swinburne Business School, BA Building, 27 John St, Hawthorn Victoria 3122, Australia, email:  
    \printead{e1} 
    \printead*{e2}
}

\thankstext{t1}{Part of this research was done while the author was a visiting fellow at the Centre for Transformative Innovation}

\end{aug}

\begin{abstract}
Variational Bayes (VB) is a recent approximate method for Bayesian inference. It has the merit of being a fast and scalable alternative to Markov Chain Monte Carlo (MCMC) but its approximation error is often unknown. In this paper, we derive the approximation error of VB in terms of mean, mode, variance, predictive density and KL divergence for the linear Gaussian multi-equation regression. Our results indicate that VB approximates the posterior mean perfectly. Factors affecting the magnitude of underestimation in the posterior variance and mode are revealed. Importantly, We demonstrate that VB estimates predictive densities accurately.
\end{abstract}



\end{frontmatter}

\section{Introduction}
Variational Bayes (VB) is an approximate method of Bayesian inference originally developed in machine learning \citep[e.g.][]{jordan1999introduction,attias2000variational}, finding its way into statistics   
\citep[e.g.][]{Ormerod2010,Blei2017} and econometrics \citep[e.g.][]{koop2018variational,gunawan2017fast}. The merit of VB is that it is fast and scalable and therefore useful in big data settings where MCMC might be difficult or infeasible (e.g. forecasting with large vector autoregressions). A drawback of the VB is that its approximation properties are still not well-understood. \\
There are a number of papers on properties of VB including but not limited to \cite{wang2005inadequacy, wang2006convergence}, \cite{hall2011asymptotic}, \cite{bickel2013asymptotic}, \cite{you2014variational}, \cite{pati2017statistical} and \cite{wang2018frequentist}. These papers focus mainly on asymptotic properties of VB. Few analytical results, if any, are available on VB's finite sample approximation error. Here we study the approximation error of VB for a linear Gaussian multiple equation regression. We focus on vector autoregressions (VAR) since large VARs are of enormous interest in macro-econometric forecasting \citep[e.g.][]{koop2013forecasting,Giannone2015} and VB is likely to be very useful in this context \citep[e.g.][]{koop2018variational} although our results are valid for any linear Gaussian multi-equation regression.\\
We derive the divergence between the mean and the mode of VB posterior and those of exact posterior, the magnitude of underestimation in posterior variance and the magnitude of the KL divergence. We show that the approximation error decreases with an increase in sample size and strength of a prior, but increases with the size of the model. We also derive the divergence in the predictive densities and demonstrate that VB estimates predictive densities well. This is important given the enormous interest in forecasting with large VARs for which MCMC is difficult or time consuming.\\
The paper is organized as follows: Section~\ref{sec:vb} provides an overview of VB while Section~\ref{sec:VAR} derives the mean-field VB posterior for both conjugate and independent prior VARs. In Section~\ref{sec:Error}, we provide the main results on approximation errors and Section~\ref{sec:Con} concludes the paper.

\section{An Overview of MFVB}\label{sec:vb} 
Let $\mathbf{y}$ be a vector of observations, $p(\mathbf{y}|\mathbf{\theta })$ the likelihood for a model and $p(\mathbf{\theta})$ a prior where the objective is to learn about the parameter vector $\mathbf{\theta }$ given the observations and the prior. Bayes theorem can be used to derive the posterior kernel easily but learning about $\mathbf{\theta}$ (its posterior density and moments) requires integration of the posterior kernel. For complex models, integration is often done through MCMC but even MCMC can sometimes become intractable (e.g. in big data settings). In VB the idea is to approximate the posterior $p(\mathbf{\theta }|\mathbf{y})$ with a density $q(\mathbf{\theta })$ that is more tractable. The optimal $q(\mathbf{\theta })$ is obtained by minimizing the KL divergence \footnote{For a more detailed review of VB see \cite{Ormerod2010} or \cite{Blei2017}.}
\[KL(q||p)=\int_{\mathbf{\theta }\in \,\Theta }{q(\mathbf{\theta })\ln [{q(\mathbf{\theta })}/{p(\mathbf{\theta }|\mathbf{y})}\;]d\mathbf{\theta }}\] over all $q(\mathbf{\theta })$s that belong to a family of densities $\Lambda .$ A commonly used family includes all $q(\mathbf{\theta })$s  that can be written as $q(\mathbf{\theta})=\prod\limits_{k=1}^{M}{{{q}_{k}}(}{{\mathbf{\theta }}_{k}})$ where $\{{{\mathbf{\theta }}_{1}},....,{{\mathbf{\theta }}_{M}}\}$ is some partition of $\mathbf{\theta }$ and the ${{q}_{k}}$s are probability density functions. This factorized form corresponds to an approximation framework known as the mean field; thus, this version of VB is often referred to as  mean-field variational Bayes (MFVB). \\
Using calculus of variations, it has been shown that the optimal ${{q}_{k}}$s are obtainable through the following iterative procedure 
\begin{align}
\begin{split}
   &q_{1}^{}{{({{\mathbf{\theta }}_{1}})}^{new}}\leftarrow \frac{\exp ({{E}_{{{\mathbf{\theta }}_{-1}}}}\log [p(\mathbf{y},\mathbf{\theta })])}{\int{\exp ({{E}_{{{\mathbf{\theta }}_{-1}}}}\log [p(\mathbf{y},\mathbf{\theta })])d{{\mathbf{\theta }}_{1}}}} \\ 
 & \quad \vdots \quad \quad \quad \quad \quad \quad \vdots  \\ 
 & \quad \vdots \quad \quad \quad \quad \quad \quad \vdots  \\ 
 & q_{M}^{}{{({{\mathbf{\theta }}_{M}})}^{new}}\leftarrow \frac{\exp ({{E}_{{{\mathbf{\theta }}_{-M}}}}\log [p(\mathbf{y},\mathbf{\theta })])}{\int{\exp ({{E}_{{{\mathbf{\theta }}_{-M}}}}\log [p(\mathbf{y},\mathbf{\theta })])d{{\mathbf{\theta }}_{M}}}}\ 
 \end{split}
 \label{system}
\end{align} 
where $p(\mathbf{y},\mathbf{\theta })=p(\mathbf{y}|\mathbf{\theta })p(\mathbf{\theta })$, ${{E}_{{{\theta }_{-k}}}}=\int{\log p(\mathbf{y},\mathbf{\theta }){{\prod\nolimits_{j\ne k}{{{q}_{j}}({{\mathbf{\theta }}_{j}})}}^{old}}}$ and ${{q}_{1}}{{({{\mathbf{\theta }}_{1}})}^{old}},....,{{q}_{M}}{{({{\mathbf{\theta }}_{M}})}^{old}}$ are current estimates for $q$. The iteration is stopped when the increase in $\ln \underline{ML}$ (defined below) becomes negligible. In practice the partition of $\mathbf{\theta }$ is often chosen in a way such that $q_{j}^{{}}({{\mathbf{\theta }}_{j}})$s are known density functions that depend on the moments of the remaining parameters. As a result, it is often sufficient to iterate over the appropriate moments extracted from the distributions in (2.1).\\
It can be shown that
\begin{equation}
\ln ML={{E}_{q}}\ln p(\mathbf{y,\theta })-{{E}_{q}}\ln q(\mathbf{\theta })+KL(q||p)
\end{equation}
where $ML$ denotes the marginal likelihood. If we define 
\begin{equation}
\ln \underline{ML}={{E}_{q}}\ln p(\mathbf{y,\theta })-{{E}_{q}}\ln q(\mathbf{\theta })
\end{equation}
then given that $KL(q||p)$ is positive, $\ln \underline{ML}$ provides a lower bound to the log-marginal-likelihood. Note that minimization of $KL(q||p)$ is equivalent to maximization of $\ln \underline{ML}$ therefore $\ln \underline{ML}$ (which can be computed using VB) is often used to monitor the convergence of (2.1). 
\section{MFVB for Vector Autoregressions}\label{sec:VAR}
Since the seminal paper of \cite{sims1980macroeconomics}, vector autoregressions have become an essential tool in macroeconomic analysis and forecasting. They are extensions of univariate autoregressive models to a vector of variables \footnote{For a detailed review of Bayesian VARs see e.g. \cite{Karlsson2013} or \cite{koop2010bayesian}.}. A VAR(d) can be written as
\begin{equation}
     {{\mathbf{y}}_{t}}={{\mathbf{\gamma }}_{0}}+{{\mathbf{\Gamma }}_{1}}{{\mathbf{y}}_{t-1}}+....+{{\mathbf{\Gamma }}_{d}}{{\mathbf{y}}_{t-d}}+{{\mathbf{\varepsilon }}_{t}}
\end{equation}
where $t=1,...,T$ indexes time, $d$ is the lag order, ${{\mathbf{y}}_{t}}$ is an $M\times 1$ vector of observations at time $t$, ${\mathbf{\gamma }}_{0}$ is an $M\times 1$ vector of constants, the ${{\mathbf{\Gamma }}_{i}}$s are $M\times M$ matrices of autoregressive coefficients and ${\mathbf{\varepsilon}}_{t}$  is an $M\times 1$ error vector with mean $\mathbf{0}$ and covariance matrix $\mathbf{\Sigma}$. 
Define $p=Md+1$ (the number of parameters in each equation), a  matrix ${{\mathbf{Y}}_{T\times M}}=({{\mathbf{y}}_{1}},...,{{\mathbf{y}}_{T}}{)}'$,   a $1\times p$ vector ${{\mathbf{x}}_{t}}=(1,{{\mathbf{{y}'}}_{T-1}},...,{{\mathbf{{y}'}}_{T-d}})$, ${{\mathbf{X}}_{T\times p}}=({{\mathbf{{x}'}}_{1}},...,{{\mathbf{{x}'}}_{T}}{)}'$, a matrix of coefficients ${{\mathbf{\Gamma }}_{p\,\,\times M}}=({{\mathbf{\gamma }}_{0}},{{\mathbf{\Gamma }}_{1}},...,{{\mathbf{\Gamma }}_{d}}{)}'$ and a matrix ${{\mathbf{E}}_{T\times M}}=({{\mathbf{\varepsilon }}_{1}},...,{{\mathbf{\varepsilon }}_{T}}{)}'$, then equation (3.1) can be written as: 
\begin{equation}
\mathbf{Y}=\mathbf{X\Gamma }+\mathbf{E}\ \ \text{with}\ \ \mathbf{E}\sim{\ }\mathcal{MN}(\mathbf{0},\mathbf{\Sigma },{{\mathbf{I}}_{T}})
\end{equation}
where $\mathcal{MN}$ denotes a matricvariate normal distribution (see Appendix-\ref{Dist}  for a definition). We first consider a natural-conjugate prior where the exact and the VB approximate posterior and the error in the MFVB approximation can be derived analytically.\\
\textbf{\textit{Normal-Wishart Conjugate Prior}}: For a VAR model, the natural-conjugate prior is given in the following normal-Wishart form:
 \begin{align}
 \begin{split}
  & p(\mathbf{\Gamma },{{\mathbf{\Sigma }}^{-1}})=p(\mathbf{\Gamma }|\mathbf{\Sigma })p({{\mathbf{\Sigma }}^{-1}})\quad  \\ 
 & p(\mathbf{\Gamma }|\mathbf{\Sigma })=\mathcal{MN}(\mathbf{\underline{\Gamma }},\mathbf{\Sigma },\mathbf{\underline{V}})\ \And p({{\mathbf{\Sigma }}^{-1}})=\mathcal{W}({{{\mathbf{\underline{S}}}}^{-1}},\underline{\upsilon }) 
 \end{split}
\end{align}
where $\mathbf{\underline{\Gamma }}$ is a $p\times M$ matrix of prior means, $\mathbf{\underline{V}}$ and $\mathbf{\underline{S}}$ are $p\times p$ and $M\times M$ positive-definite matrices, $\mathcal{W}$ and $\underline{\upsilon }$ denote the Wishart distribution and its degrees of freedom. In this case, it is well-known that the exact posterior is a matricvariate normal-Wishart distribution \citep[e.g.][]{Karlsson2013}:
\begin{align}
\begin{split}
  & p(\mathbf{\Gamma }|{{\mathbf{\Sigma }}^{-1}},\mathbf{Y})=\mathcal{MN}(\mathbf{\bar{\Gamma }},{{\mathbf{\Sigma }}^{{}}},\mathbf{\bar{V}})\  \\ 
 & \ \ \ \mathbf{\bar{V}=}{{[{{{\mathbf{\underline{V}}}}^{-1}}\mathbf{+{X}'X}]}^{-1}}\ \And \ \mathbf{\bar{\Gamma }=\bar{V}}[{{{\mathbf{\underline{V}}}}^{-1}}\mathbf{\underline{\Gamma }+{X}'Y}]\quad  \\ 
 & p({{\mathbf{\Sigma }}^{-1}}|\mathbf{Y})=\mathcal{W}({{{\mathbf{\bar{S}}}}^{-1}},\bar{\upsilon })\ \ \text{where}\ \ \bar{\upsilon }=T+\underline{\upsilon } \\ 
 & \text{  }\ \mathbf{\bar{S}=}(\mathbf{Y}-\mathbf{X\bar{\Gamma }}{)}'(\mathbf{Y}-\mathbf{X\bar{\Gamma }})\mathbf{+\underline{S}+}(\mathbf{\bar{\Gamma }}-\mathbf{\underline{\Gamma }}{)}'{{{\mathbf{\underline{V}}}}^{-1}}(\mathbf{\bar{\Gamma }}-\mathbf{\underline{\Gamma }})\  \\ 
 \end{split}
\end{align}
Other quantities of interest are marginal posteriors, posterior moments, marginal likelihoods and predictive densities that are available in analytical forms and are given in Table-\ref{Tab1}.\\
Now let’s consider MFVB inference where the VB approximate posterior density is factorized into $q(\mathbf{\Gamma },{{\mathbf{\Sigma }}^{-1}})=q(\mathbf{\Gamma })q({{\mathbf{\Sigma }}^{-1}})$. 
Using (2.1) it can be shown (see the Appendix-\ref{3.5}) that the VB posterior is
\begin{align}
\begin{split}
  & q(\mathbf{\Gamma })=\mathcal{MN}\{\mathbf{\bar{\Gamma }},{{\overline{{{\mathbf{\Sigma }}^{-1}}}}^{\,-1}},\mathbf{\bar{V}}\} \\ 
 & \quad \mathbf{\bar{V}=}{{[{{{\mathbf{\underline{V}}}}^{-1}}\mathbf{+{X}'X}]}^{-1}}\ \ \ \mathbf{\bar{\Gamma }=\bar{V}}[{{{\mathbf{\underline{V}}}}^{-1}}\mathbf{\underline{\Gamma }+{X}'Y}] \\ 
 & \quad \overline{{{\mathbf{\Sigma }}^{-1}}}=\bar{\upsilon }{{{\mathbf{\bar{S}}}}^{-1}} \\ 
 & \quad \mathbf{\bar{S}=}(\mathbf{Y}-\mathbf{X\bar{\Gamma }}{)}'(\mathbf{Y}-\mathbf{X\bar{\Gamma }})\mathbf{+\underline{S}+}(\mathbf{\bar{\Gamma }}-\mathbf{\underline{\Gamma }}{)}'{{{\mathbf{\underline{V}}}}^{-1}}(\mathbf{\bar{\Gamma }}-\mathbf{\underline{\Gamma }})\  \\ 
 & q({{\mathbf{\Sigma }}^{-1}})=\mathcal{W}({{{\mathbf{\bar{S}}}}_{q}}^{-1},{{{\bar{\upsilon }}}_{q}})\quad \text{ } \\ 
 & \quad {{{\mathbf{\bar{S}}}}_{q}}\mathbf{=}\frac{{{{\bar{\upsilon }}}_{q}}}{{\bar{\upsilon }}}\mathbf{\bar{S}}\ \ \ \ {{{\bar{\upsilon }}}_{q}}=T+p+\underline{\upsilon }\quad \  
 \end{split}
\end{align} 
without a need for an iterative process !. Related quantities of interest such as marginal posteriors, posterior moments, marginal likelihoods and predictive densities are given in Table-\ref{Tab1}.\\
\textbf{\textit{Normal-Wishart Independent Prior}}: A conjugate prior has the advantage that one can derive the approximation errors analytically. The potential drawback is that the results might be crucially dependent on the conjugacy assumption. Therefore, we also consider a VAR with an independent prior. In this case we cannot derive the approximation error analytically, but we use a real data example to see the extent to which the results differ from the conjugate VAR’s results.\\
To discuss VAR with an independent prior we appeal to another way of writing a VAR model \citep[see e.g.][]{koop2010bayesian}:\\
\begin{align*}
\left( \begin{aligned} 
  & {{\mathbf{y}}_{1}} \\ 
 & \ \vdots  \\ 
 & {{\mathbf{y}}_{T}} \\ 
\end{aligned} \right)
=
\left( \begin{aligned}
  & {{\mathbf{Z}}_{1}} \\ 
 & \ \vdots  \\ 
 & {{\mathbf{Z}}_{T}} \\ 
 \end{aligned} \right)
\left( \begin{aligned}
  & {{\mathbf{\beta }}_{\mathbf{1}}} \\ 
 & \ \vdots  \\ 
 & {{\mathbf{\beta }}_{M}} \\ 
\end{aligned} \right)
+\left( \begin{aligned}
  & {{\mathbf{\varepsilon }}_{\mathbf{1}}} \\ 
 & \ \vdots  \\ 
 & {{\mathbf{\varepsilon }}_{M}} \\ 
\end{aligned} \right)
\end{align*}
 $\And \ {{\mathbf{Z}}_{t}}=\left( \begin{matrix}
   {{{\mathbf{{z}'}}}_{1t}} & \mathbf{0} & \cdots  & \mathbf{0}  \\
   \mathbf{0} & {{{\mathbf{{z}'}}}_{2t}} & \ddots  & \vdots   \\
   \vdots  & \ddots  & \ddots  & \mathbf{0}  \\
   \mathbf{0} & \cdots  & \mathbf{0} & {{{\mathbf{{z}'}}}_{M\,t}}  \\
\end{matrix} \right)$
where if $\mathbf{z'}_{mt}=\mathbf{x}_t$ for $m=1,...,M$, we obtain the same equation as (3.1). In a more compact form:
 \begin{equation}
 \mathbf{y}=\mathbf{Z\beta }+\mathbf{\varepsilon } \text{with}  \mathbf{\varepsilon }\sim \mathcal{N}(\mathbf{0},{{\mathbf{I}}_{T}}\otimes {{\mathbf{\Sigma }}_{M}})
\end{equation}
\, where $\mathcal{N}$ denotes a multivariate normal distribution. For the independent case, we assume that the parameter vector $\mathbf{\beta }$ and the precision matrix $\mathbf{\Sigma }_{{}}^{-1}$ are a priori independent:
  \begin{align}
  \begin{split}
  & \quad \quad \quad  p(\mathbf{\beta },{{\mathbf{\Sigma }}^{-1}})=p(\mathbf{\beta })p({{\mathbf{\Sigma }}^{-1}})\  \\ 
 & \text{ with} \quad p(\mathbf{\beta })=\mathcal{N}(\mathbf{\underline{\beta }},\mathbf{\underline{V}}),\ p({{\mathbf{\Sigma }}^{-1}})=\mathcal{W}({{{\mathbf{\underline{S}}}}^{-1}},\underline{\upsilon })
 \end{split}
\end{align}
It is well-known that under this prior, the posterior can be obtained through a Gibbs sampler \citep[e.g.][]{Karlsson2013} defined by
\begin{align}
\begin{split}
  & p(\mathbf{\beta }|{{\mathbf{\Sigma }}^{-1}},\mathbf{y})=\mathcal{N}(\mathbf{\bar{\beta }},\mathbf{\bar{V}})\quad \quad \ \  \\ 
 & \ \ \mathbf{\bar{V}}\text{=}[{{{\mathbf{\underline{V}}}}^{-1}}+\sum\limits_{t=1}^{T}{{{{\mathbf{{Z}'}}}_{t}}}{{\mathbf{\Sigma }}^{-1}}{{\mathbf{Z}}_{t}}]^{-1},\ \ \mathbf{\bar{\beta }}=\mathbf{\bar{V}}[{{{\mathbf{\underline{V}}}}^{-1}}\mathbf{\underline{\beta }}+\sum\limits_{t=1}^{T}{{{{\mathbf{{Z}'}}}_{t}}}{{\mathbf{\Sigma }}^{-1}}{{\mathbf{y}}_{t}}] \\ 
 & p({{\mathbf{\Sigma }}^{-1}}|\mathbf{\beta },\mathbf{y})=\mathcal{W}({{{\mathbf{\bar{S}}}}^{-1}},\bar{\upsilon })\quad \quad \bar{\upsilon }=T+\underline{\upsilon } \\ 
 & \ \ \ \mathbf{\bar{S}}\text{=}\mathbf{\underline{S}}+\sum\limits_{t=1}^{T}{({{\mathbf{y}}_{t}}-{{\mathbf{Z}}_{t}}\mathbf{\beta })({{\mathbf{y}}_{t}}-{{\mathbf{Z}}_{t}}\mathbf{\beta }{)}'}\quad  \
 \end{split}
\end{align} 
Marginal likelihoods and predictive densities are not available in analytical form in this case, but can be obtained through simulation \citep[e.g.][]{Karlsson2013}.\\
\\
 Now let's assume $q(\mathbf{\beta },{{\mathbf{\Sigma }}^{-1}})=q(\mathbf{\beta})q({{\mathbf{\Sigma }}^{-1}})$, then MFVB posterior can be obtained using (2.1) as 
\begin{align}
\begin{split}
 & q(\mathbf{\beta })=\mathcal{N}({{{\mathbf{\bar{\beta }}}}_{q}},{{{\mathbf{\bar{V}}}}_{q}}) \\ 
 & q({{\mathbf{\Sigma }}^{-1}})=\mathcal{W}(\mathbf{\bar{S}}_{q}^{-1},\bar{\upsilon })\quad \quad \bar{\upsilon }=T+\underline{\upsilon } \end{split} 
\end{align}                                
where $\{{{\mathbf{\bar{\beta}}}_{q}},{{\mathbf{\bar{V}}}_{q}},\mathbf{\bar{S}}_{q}^{-1}\}$ are obtained by using a starting value for $\overline{\mathbf{\Sigma }_{q}^{-1}}$ and iterating the following until convergence
\begin{align*}
\begin{split}
  & {{{\mathbf{\bar{V}}}}_{q}}\text{=}[{{{\mathbf{\underline{V}}}}^{-1}}+\sum\limits_{t=1}^{T}{{{{\mathbf{{Z}'}}}_{t}}}\overline{\mathbf{\Sigma }_{q}^{-1}}{{\mathbf{Z}}_{t}}]^{-1},\ \ {{{\mathbf{\bar{\beta }}}}_{q}}=\mathbf{\bar{V}}[{{{\mathbf{\underline{V}}}}^{-1}}\mathbf{\underline{\beta }}+\sum\limits_{t=1}^{T}{{{{\mathbf{{Z}'}}}_{t}}}\overline{\mathbf{\Sigma }_{q}^{-1}}{{\mathbf{y}}_{t}}] \\ 
 & \ {{{\mathbf{\bar{S}}}}_{q}}\text{=}\mathbf{\underline{S}}+\sum\limits_{t=1}^{T}{({{\mathbf{y}}_{t}}-{{\mathbf{Z}}_{t}}{{{\mathbf{\bar{\beta }}}}_{q}})({{\mathbf{y}}_{t}}-{{\mathbf{Z}}_{t}}{{{\mathbf{\bar{\beta }}}}_{q}}{)}'}+{{\mathbf{Z}}_{t}}{{{\mathbf{\bar{V}}}}_{q}}{{{\mathbf{{Z}'}}}_{t}},\ \overline{\mathbf{\Sigma }_{q}^{-1}}=\bar{\upsilon }\mathbf{\bar{S}}_{q}^{-1} \
 \end{split}
 \end{align*} 
$\ln \underline{ML}={{E}_{q}}\ln p(\mathbf{y,\theta })-{{E}_{q}}\ln q(\mathbf{\theta })$, which is used for monitoring of the convergence, can be obtained using (2.3), (A.1), and the entropy of normal and Wishart distributions, as
\begin{align*}
\begin{split}
  & \ln \underline{ML}=\frac{p}{2}-\frac{MT}{2}\ln \pi +\ln {{\Gamma }_{M}}\left( \frac{{\bar{\upsilon }}}{2} \right)-\ln {{\Gamma }_{M}}\left( \frac{{\underline{\upsilon }}}{2} \right)+\frac{\ln \left| {{{\mathbf{\bar{V}}}}_{q}} \right|-\ln |\mathbf{\underline{V}}|}{2} \\ 
 & \quad +\frac{\bar{\upsilon }\ln \left| \mathbf{\bar{S}}_{q}^{-1} \right|-\underline{\upsilon }\ln \left| {{{\mathbf{\underline{S}}}}^{-1}} \right|}{2}-\frac{tr\left\{ {{{\mathbf{\underline{V}}}}^{-1}}[({{{\mathbf{\bar{\beta }}}}_{q}}-\mathbf{\underline{\beta }})({{{\mathbf{\bar{\beta }}}}_{q}}-\mathbf{\underline{\beta }}{)}'+{{{\mathbf{\bar{V}}}}_{q}}] \right\}}{2}\quad 
 \end{split}
\end{align*}
where ${\Gamma }_{M}$ is the multivariate gamma function, $\left| \ .\  \right|$ is the determinant and $tr\{.\}$ denotes the trace of a matrix.\\
While calculating the exact predictive density requires simulation, using arguments similar to those used in Appendix-\ref{Der}, the MFVB predictive density can be obtained as sum of a normal and a t-distribution with the following mean and variance
\begin{align}
\begin{split}
 &{{E}_{q}}({{\mathbf{y}}_{T+1}}|\mathbf{y})={{E}_{q}}[\mathbf{Z}_{T+1}^{{}}\mathbf{\beta }|\mathbf{y}]+{{E}_{q}}({{\mathbf{\varepsilon }}_{T+1}}|\mathbf{y})=\mathbf{Z}_{T+1}^{{}}{{\mathbf{\bar{\beta }}}_{q}}\\
  & Va{{r}_{q}}({{\mathbf{y}}_{T+1}}|\mathbf{Y})=Va{{r}_{q}}[\mathbf{Z}_{T+1}^{{}}{{\mathbf{\beta }}_{q}}|\mathbf{y}]+Va{{r}_{q}}({{\mathbf{\varepsilon }}_{T+1}}|\mathbf{y}) \\ 
 &  =\mathbf{Z}_{T+1}^{{}}{{{\mathbf{\bar{V}}}}_{q}}\mathbf{{Z}'}_{T+1}^{{}}+\frac{{{{\mathbf{\bar{S}}}}_{q}}}{(\bar{\upsilon }-2)} \
 \end{split}
\end{align}
  \begin{table} [h!]
   \begin{adjustwidth}{-2.2cm}{-1.5cm}
\renewcommand{\arraystretch}{1.5}
    \begin{threeparttable}
\centering 
\caption{Comparing various features of the exact and MFVB posteriors for natural conjugate VARs}
\label{Tab1}
\begin{tabular}{|l|l|l|c | c | c|} 
 \hline
  &	Exact Posterior &	MFVB Posterior \\
  \hline
\shortstack{Marginal \\ Posterior} &	
$\begin{array}{l}
   p(\mathbf{\Gamma }|\mathbf{Y})=\mathcal{MT}(\mathbf{\bar{\Gamma }},\mathbf{\bar{S}},\mathbf{\bar{V}},\bar{\upsilon }) \\ p({{\mathbf{\Sigma }}^{-1}}|\mathbf{Y})=\mathcal{W}({{{\mathbf{\bar{S}}}}^{-1}},\bar{\upsilon })\quad \bar{\upsilon }=T+\underline{\upsilon } \\ 
\end{array}$
& 
$\begin{array}{l}
   q(\mathbf{\Gamma })=\mathcal{MN}\{\mathbf{\bar{\Gamma }},{{\overline{{{\mathbf{\Sigma }}^{-1}}}}^{-1}},\mathbf{\bar{V}}\} \\ 
  q({{\mathbf{\Sigma }}^{-1}})=\mathcal{W}({{{\mathbf{\bar{S}}}}_{q}}^{-1},{{{\bar{\upsilon }}}_{q}})\quad {{{\bar{\upsilon }}}_{q}}=T+\underline{\upsilon }+p \\ \end{array}$\\
 \hline
 \shortstack{Posterior\\Mean} & 
 $\begin{array}{l}
  {{E}_{p}}(\mathbf{\Gamma })=\mathbf{\bar{\Gamma }} \\ 
  {{E}_{p}}({{\mathbf{\Sigma }}^{-1}})=\bar{\upsilon }{{{\mathbf{\bar{S}}}}^{-1}} \\ 
\end{array}$ &
$\begin{array}{l}
{{E}_{q}}(\mathbf{\Gamma })=\mathbf{\bar{\Gamma }} \\
 {{E}_{q}}(\mathbf{\Sigma }_{{}}^{-1})={{\bar{\upsilon }}_{q}}{{\mathbf{\bar{S}}}_{q}}^{-1}=\bar{\upsilon }{{\mathbf{\bar{S}}}^{-1}}
 \end{array}$\\
\hline
\shortstack{Posterior\\Mode} & 
 $\begin{array}{l}
  Mod{{e}_{p}}(\mathbf{\Gamma })=\mathbf{\bar{\Gamma }} \\ 
  Mod{{e}_{p}}({{\mathbf{\Sigma }}^{-1}})=(\bar{\upsilon }+p-m-1){{{\mathbf{\bar{S}}}}^{-1}} \\
\end{array}$ &
$\begin{array}{l}
Mod{{e}_{q}}(\mathbf{\Gamma })=\mathbf{\bar{\Gamma }} \\ 
Mod{{e}_{q}}({{\mathbf{\Sigma }}^{-1}})=\dfrac{{\bar{\upsilon }}}{{{{\bar{\upsilon }}}_{q}}}(\bar{\upsilon }+p-m-1){{{\mathbf{\bar{S}}}}^{-1}} \\ 
 \end{array}$\\
\hline
\shortstack{Posterior\\Variance} & 
 $\begin{array}{l}
  Va{{r}_{p}}(\mathbf{\gamma })=\dfrac{1}{\bar{\upsilon }-M-1}\mathbf{\bar{S}}\otimes \mathbf{\bar{V}} \\ 
  Va{{r}_{p}}(\sigma _{ij}^{*})=\bar{\upsilon }\left( s{{_{ij}^{*}}^{2}}+s_{ii}^{*}s_{jj}^{*} \right) \\
  \text{where}\,\ {{\mathbf{\Sigma }}^{-1}}=\{\sigma _{ij}^{*}\}  \,\,\,\,\ {{\mathbf{\bar{S}}}^{-1}}=\{s_{ij}^{*}\}\\
\end{array}$ &
$\begin{array}{l}
Va{{r}_{q}}(\mathbf{\gamma })=\dfrac{1}{{\bar{\upsilon }}}\mathbf{\bar{S}}\otimes \mathbf{\bar{V}}\\
Va{{r}_{q}}(\sigma _{ij}^{*})=\dfrac{\bar{\upsilon }_{{}}^{2}}{\bar{\upsilon }_{q}^{{}}}\left( s{{_{ij}^{*}}^{2}}+s_{ii}^{*}s_{jj}^{*} \right) \\ 
\text{where}\,\ \mathbf{\Sigma }_{{}}^{-1}=\{\sigma _{ij}^{*}\} \,\,\,\,\ {{\mathbf{\bar{S}}}^{-1}}=\{s_{ij}^{*}\}\ \\ 
 \end{array}$\\
\hline
\shortstack{\\Log of \\ Marginal\\Likelihood} & 
 $\begin{array}{l}
-\dfrac{MT}{2}\ln \pi +\dfrac{M}{2}(\ln \left| {\mathbf{\bar{V}}} \right|-\ln |\mathbf{\underline{V}}|)-\dfrac{{\bar{\upsilon }}}{2}\ln \left| {\mathbf{\bar{S}}} \right| \\ 
 +\dfrac{{\underline{\upsilon }}}{2}\ln \left| {\mathbf{\underline{S}}} \right|+\ln {{\Gamma }_{M}}\left( {\dfrac{\bar{\upsilon }}{2}}\; \right)-\ln {{\Gamma }_{M}}({\dfrac{\underline{\upsilon }}{2}}\;) \\
\end{array}$ &
$\begin{array}{l}
-\dfrac{MT}{2}\ln \pi +\dfrac{M}{2}\left( \ln \left| {\mathbf{\bar{V}}} \right|-\ln |\mathbf{\underline{V}}| \right)+\dfrac{{\bar{\upsilon }}}{2}\ln \left| {{{\mathbf{\bar{S}}}}^{-1}} \right|+\dfrac{{\underline{\upsilon }}}{2}\ln \left| {\mathbf{\underline{S}}} \right| \\ 
 -\dfrac{Mp}{2}\ln 2e+\dfrac{M}{2}\ln \left( \dfrac{{{{\bar{\upsilon }}}_{{}}}^{{{{\bar{\upsilon }}}_{{}}}}}{{{{\bar{\upsilon }}}_{q}}^{{{{\bar{\upsilon }}}_{q}}}} \right)+\ln {{\Gamma }_{M}}\left( {\dfrac{{{\bar{\upsilon }}}_{q}}{2}}\; \right)-\ln {{\Gamma }_{M}}({\dfrac{\underline{\upsilon }}{2}}\;) \\
 \end{array}$\\
\hline
\shortstack{Predictive\\Density} & 
 $\begin{array}{l}
 \text{t-distribution with} \\
  {{E}_{p}}({{\mathbf{y}}_{T+1}}|\mathbf{Y})=({{\mathbf{x}}_{T+1}}\mathbf{\bar{\Gamma }}{)}' \\ 
 Va{{r}_{p}}({{\mathbf{y}}_{T+1}}|\mathbf{Y})=(1+{{\mathbf{x}}_{T+1}}\mathbf{\bar{V}}{{{\mathbf{{x}'}}}_{T+1}})\dfrac{\,\mathbf{\bar{S}}}{\bar{\upsilon }-2} \\
\end{array}$ &
$\begin{array}{l}
\text{Sum of a normal and a t-distribution with} \\
 {{E}_{q}}({{\mathbf{y}}_{T+1}}|\mathbf{Y})=({{\mathbf{x}}_{T+1}}\mathbf{\bar{\Gamma }}{)}' \\ 
 Va{{r}_{q}}({{\mathbf{y}}_{T+1}}|\mathbf{Y})=\left( \dfrac{{{{\bar{\upsilon }}}_{q}}}{{{{\bar{\upsilon }}}_{q}}-2}+\mathbf{x}_{T+1}^{{}}\mathbf{\bar{V}{x}'}_{T+1}^{{}} \right)\dfrac{{\mathbf{\bar{S}}}}{{\bar{\upsilon }}} \\ 
 \end{array}$\\
\hline
\end{tabular}
\begin{tablenotes}
\item \it The results for the Exact Posterior are taken from the existing literature. The results for MFVB are new but not difficult to derive given (3.5).  Modes for the exact posterior are derived by maximizing the posterior and for MFVB using the known results on the modes of Normal and Wishart distributions. $\mathcal{MT}$ denotes the matricvariate t-student distribution, $\mathbf{\gamma }=vec(\mathbf{\Gamma })$, $\left| \ .\  \right|$ denotes the determinant and ${{\Gamma }_{M}}(.)$ is the multivariate gamma function. See Appendix-\ref{3.5}, Appendix-\ref{Dist} and \cite{Karlsson2013} for further information.
\end{tablenotes}
    \end{threeparttable}
\end{adjustwidth}
\end{table}
\section{MFVB Approximation Error}\label{sec:Error}
Table-\ref{Tab1} compares various features of MFVB and of the exact posterior for the conjugate prior case. As this table shows, the posterior means for both $\mathbf{\Gamma }$ and ${{\mathbf{\Sigma }}^{-1}}$ are equal but MFVB underestimates variances of the parameters, confirming the folklore. The variance of ${{\mathbf{\Gamma }}_{q}}$ is smaller than the variance of $\mathbf{\Gamma }$ by a factor of $1-\frac{M+1}{T+\underline{\upsilon }}$ which is close to one unless the number of equations is large (see Table-\ref{Tab2}). Variances of elements of  $\mathbf{\Sigma }_{q}^{-1}$  are smaller by a factor of $1-({p+1)}/{(T+\underline{\upsilon })}\;$ indicating small underestimation when $p$ (the number of parameters in each equation) is small compared to $T+\underline{\upsilon }$. However, it could be substantial for large VARs (see Table-\ref{Tab2}). The mode of the approximate posterior is equal to the mode of the exact posterior for $\mathbf{\Gamma }$ but for $\mathbf{\Sigma }_{q}^{-1}$ it is smaller by a factor of $1-{p}/{(T+\underline{\upsilon }+p)}\;$.  
The KL divergence which is a measure of closeness of the approximate posterior to the exact posterior can be derived as
\[KL=\ln ML-\ln \underline{ML}=-\dfrac{Mp}{2}(\ln 2+1)+\dfrac{M}{2}\ln \left( \dfrac{{{{\bar{\upsilon }}}_{q}}^{{{{\bar{\upsilon }}}_{q}}}}{{{{\bar{\upsilon }}}_{{}}}^{{{{\bar{\upsilon }}}_{{}}}}} \right)-\ln \left( \dfrac{{{\Gamma }_{M}}({{{{\bar{\upsilon }}}_{q}}}/{2}\;)}{{{\Gamma }_{M}}\left( {{{{\bar{\upsilon }}}_{{}}}}/{2}\; \right)} \right)\] 
and using Stirling’s approximation to gamma functions as \[KL\sim \dfrac{1}{2}\sum\limits_{j=1}^{M}{\ln \left( \dfrac{{{\left( \bar{\upsilon }-j+1 \right)}^{(\bar{\upsilon }-j)}}{{{\bar{\upsilon }}}_{q}}^{{{{\bar{\upsilon }}}_{q}}}}{{{\left( {{{\bar{\upsilon }}}_{q}}-j+1 \right)}^{({{{\bar{\upsilon }}}_{q}}-j)}}{{{\bar{\upsilon }}}_{{}}}^{{\bar{\upsilon }}}} \right)}\]
Note that KL increases as the size of the model (i.e. the numbers of parameters in each equation) increases and decreases with an increase in the number of observations $T$, and with the strength of the prior for ${{\mathbf{\Sigma }}^{-1}}$ (i.e. $\underline{\upsilon }$). The KL does not depend on the data and the other hyper-parameters !\\
Note that
\begin{align*}
  & T\to \infty \Rightarrow \frac{Var_q(\sigma _{ij}^{*})}{Var_p(\sigma _{ij}^{*})}\to 1,\,\ \frac{Var_q(\mathbf{\gamma }_h)}{Var_p(\mathbf{\gamma}_h)}\to {1},KL\to 0 \\ 
 & \underline{\upsilon }\to \infty \Rightarrow \frac{Var_q(\sigma _{ij}^{*})}{Var_p(\sigma _{ij}^{*})}\to 1,\ \frac{Var_q(\mathbf{\gamma }_h)}{Var_p(\mathbf{\gamma}_h)}\to {1},KL\to 0 \\ 
 & p\to \infty \Rightarrow \frac{Var_q(\sigma _{ij}^{*})}{Var_p(\sigma _{ij}^{*})}\to 0,\ KL\to \infty  \\ 
\end{align*}
where $i,j=1,..,M$ and $h=1,...,Mp$. The first line shows that with an increase in the number of observations, variances of parameters based on MFVB approaches variances of parameters from the exact Bayes. The second line shows the same results with an increase in strength of the prior for $\Sigma^{-1}$.    \\Also note that the predictive densities based on the exact and MFVB posteriors share the same mean and an almost equal variance. The size of the model does not seem to affect predictive densities (see Tables \ref{Tab1} and \ref{Tab2}).\\
To study the approximation error for the VAR with an independent prior, we estimate a small (M=3) and a medium VAR (with M=7) using the same data set as in \cite{Giannone2015} assuming four lags (d = 4). The data set is based on the benchmark US data set of \cite{Stock2008} and consists of 200 observations on US macro quarterly variables from the first quarter of 1959 to the last quarter of 2008. The seven variables are real GDP, federal funds rate, GDP deflator, real investment, real consumption, hours worked and real compensation per hour. Hyper-parameters (i.e. $\mathbf{\underline{\Gamma }},\mathbf{\underline{S}},\mathbf{\underline{V}},\underline{\upsilon },\mathbf{\underline{\beta }}$) have been chosen based on the so called Minnesota prior as described in e.g. \cite{Karlsson2013}. \\

\begin{table} [h!]
\renewcommand{\arraystretch}{1.5}
    \begin{threeparttable}
\centering 
\caption{Comparisons for conjugate VARs of different sizes}
\label{Tab2}
\begin{tabular}{c| c c c c}
\hline\hline 
\it {VAR} & $\dfrac{Var_q(\mathbf{\Gamma})}{Var_p(\mathbf{\Gamma})}$ & $\dfrac{Var_q(\sigma _{ij}^*)}{{Var_p}(\sigma _{ij}^*)}$ &	$\dfrac{Mode_q(\sigma _{ij}^*)}{Mode_p(\sigma _{ij}^*)}$ & $\dfrac{Var_q(\mathbf{y}_{T+1}|\mathbf{Y})}{Var_p(\mathbf{y}_{T+1}|\mathbf{Y})}$\\
\hline\hline
Small M=3      & 0.980  & 0.930  & 0.939  &  0.990   \\
Medium M=7     & 0.961  & 0.853  & 0.876  &  0.990    \\
Large M=20     & 0.903	& 0.622	 & 0.728  &  0.991    \\
\hline
\end{tabular} 
\begin{tablenotes}
\item \it $\sigma _{ij}^{*}$s are elements of ${{\mathbf{\Sigma }}^{-1}}$. Note that for large VARs, variance underestimation for the parameters is the substantial but the variance of predictive density is not affected. The posterior and predictive means are equal in all cases.  Results are based on T = 196 and $\underline{\upsilon}=M+2$.
\end{tablenotes}
    \end{threeparttable}
\end{table}

\begin{table} [h!]
\renewcommand{\arraystretch}{1.5}
    \begin{threeparttable}
\centering 
\caption{KL for conjugate and independent VARs}
\label{Tab3}
\begin{tabular}{c c |c c c}
\hline\hline 
  Model & Size &	$\ln ML$ &	$\ln \underline{ML}$ &	\it KL \\
\hline\hline
 Conjugate VAR & Small &	1370.62 &	1370.43 &	0.189 \\
Conjugate VAR &	Medium &	2910.09 &	2908.22 &	1.874\\
\hline
Independent VAR &  Small & 1368.80 &	1368.61	& 0.193 \\
Independent VAR &	Medium & 	2882.12 &	2880.38  &	1.742 \\
\hline
\end{tabular} 
\begin{tablenotes}
\item \it The result for the  conjugate case has been derived using Table-\ref{Tab1}, $\ln \underline{ML}$ for the independent case has been derived using equation (3.10) and for $\ln ML$ using an accurate method described in \cite{hajargasht2018accurate} based on 35000 Gibbs draws with 5000 as the burn-in. Simulation errors were quite small compared to the values and have little impacts on the results.
\end{tablenotes}
    \end{threeparttable}
\end{table}
\noindent Table-\ref{Tab3} reports $\ln ML$, $\ln \underline{ML}$  and KL divergence for both the conjugate and independent prior models. As it can be seen, the KL divergences for conjugate and independent VARs, although not equal,  are of the same order of magnitude and they both increase with the size of the model. 
Our other results (not reported here) show that KL divergence decreases with increasing   $\underline{\upsilon }$ and $T$. \\
Since the number of variables in VAR models is large, we cannot present comparisons for each of the parameters here, but Table-\ref{Tab4} gives means and variances of the diagonal elements of the precision matrix for the small VAR.  The results suggest that MFVB's mean of  ${{\mathbf{\Sigma }}^{-1}}$  is very well estimated but its variance is underestimated and the amount of underestimation is of a similar order to that from the the conjugate case (see Table-\ref{Tab6} for the medium VAR’s estimation results).

\begin{table} [h!]
\renewcommand{\arraystretch}{1.2}
    \begin{threeparttable}
\centering 
\caption{Comparing posterior mean and \,\ variance of some parameters for small VAR}
\label{Tab4}
\begin{tabular}{|c c| c c|}
\hline
\multicolumn{2}{|c|}{Conjugate VAR} & \multicolumn{2}{c|}{Independent VAR}
\\
\hline
{$\dfrac{{{E}_{q}}(\sigma _{ii}^{*})}{{{E}_{p}}(\sigma _{ii}^{*})}$} &	{$\dfrac{{Var_q}(\sigma _{ii}^{*})}{{Var_p}(\sigma _{ii}^{*})}$} &	{$\dfrac{{{E}_{q}}(\sigma _{ii}^{*})}{{{E}_{p}}(\sigma _{ii}^{*})}$} &	{$\dfrac{{Var_q}(\sigma _{ii}^{*})}{{Var_p}(\sigma _{ii}^{*})}$} \\
 \hline
1 &	0.939 &	1.001 &	0.941 \\
1 &	0.939 &	1.002 &	0.947 \\
1 &	0.939 &	0.999 &	0.927 \\
\hline
\end{tabular} 
\begin{tablenotes}
\item \it $\sigma _{ii}^{*}$s are diagonal elements of ${{\mathbf{\Sigma }}^{-1}}$. The estimates for the independent case are based on 35000 Gibbs draws with 5000 as burn-in. Here q refers to VB posterior and p refers to exact or MCMC posterior. 
\end{tablenotes}
    \end{threeparttable}
\end{table}
\noindent Table-\ref{Tab5} gives the ratios of means and variances from VB to those from the exact (or MCMC) predictive densities for the small VAR [see Table-\ref{Tab6} for medium VAR]. For the conjugate case, the means are equal but the variances are smaller for VB, although the difference is negligible. A similar statement can be made for the independent prior case although  underestimation seems slightly higher (perhaps more a result of simulation errors).
\begin{table} [h!]
\renewcommand{\arraystretch}{1.2}
    \begin{threeparttable}
\centering 
\caption{Predictive density’s mean and variance }
\label{Tab5}
\begin{tabular}{|c c| c c|}
\hline
\multicolumn{2}{|c|}{Conjugate VAR} & \multicolumn{2}{c|}{Independent VAR}
\\
\hline
$\dfrac{{{E}_{q}}({{\mathbf{y}}_{T+1}}|\mathbf{Y})}{{{E}_{p}}({{\mathbf{y}}_{T+1}}|\mathbf{Y})}$ &	$\dfrac{{Var_{q}}({{\mathbf{y}}_{T+1}}|\mathbf{Y})}{{Var_{p}}({{\mathbf{y}}_{T+1}}|\mathbf{Y})}$ &	$\dfrac{{{E}_{q}}({{\mathbf{y}}_{T+1}}|\mathbf{y})}{{{E}_{p}}({{\mathbf{y}}_{T+1}}|\mathbf{y})}$ &	$\dfrac{{Var_{q}}({{\mathbf{y}}_{T+1}}|\mathbf{y})}{{Var_{p}}({{\mathbf{y}}_{T+1}}|\mathbf{y})}$ \\
\hline
1 &	0.9987 &	1.0000 &	0.9923 \\
1 &	0.9987 &	1.0000 &	0.9874 \\
1 &	0.9987 &	0.9980 &	0.9869 \\
\hline
\end{tabular} 
\begin{tablenotes}
\item \it The estimates for the VAR with independent prior is based on 35000 Gibbs draws with 5000 as burn-in period. 
\end{tablenotes}
    \end{threeparttable}
\end{table}

\begin{table} [h!]
\renewcommand{\arraystretch}{1.2}
    \begin{threeparttable}
\centering 
\caption{Mean and variance ratios of  MFVB  to MCMC for medium VAR (M=7) with the independent prior}
\label{Tab6}
\begin{tabular}{|c c| c c|}
\hline
\multicolumn{2}{|c|}{Variances} & \multicolumn{2}{c|}{Predictive Density}
\\
\hline
	{$\dfrac{{{E}_{q}}(\sigma _{ii}^{*})}{{{E}_{p}}(\sigma _{ii}^{*})}$} &	{$\dfrac{{Var_{q}}(\sigma _{ii}^{*})}{{Var_{p}}(\sigma _{ii}^{*})}$} &	{$\dfrac{{{E}_{q}}({{\mathbf{y}}_{T+1}}|\mathbf{y})}{{{E}_{p}}({{\mathbf{y}}_{T+1}}|\mathbf{y})}$} &	{$\dfrac{{Var_{q}}({{\mathbf{y}}_{T+1}}|\mathbf{y})}{{Var_{p}}({{\mathbf{y}}_{T+1}}|\mathbf{y})}$} \\
	\hline
	0.9993 &	0.8809 &	1.0000 &	0.9936 \\
	0.9984 &	0.8811 &	1.0000 &	0.9919 \\
	0.9984 &	0.8776 &	1.0173 &	0.9930 \\
	0.9991 &	0.8819 &	1.0000 &	0.9911 \\
	0.9989 &	0.8710 &	0.9998 &	0.9827 \\
	0.9993 &	0.9029 &	1.0000 &	0.9966 \\
	0.9987 &	0.9217 &	1.0000 &	0.9887 \\
\hline
\end{tabular} 
\begin{tablenotes}
\item \it These results are based on 105000 Gibbs draws with 5000 as burn-in. These numbers support our claim that results similar to those of the conjugate prior VARs hold for the independent prior case.
\end{tablenotes}
    \end{threeparttable}
\end{table}
Tables \ref{Tab3} to \ref{Tab6} suggest VB's approximation errors derived for the conjugate prior case also holds for the independent prior case although not in an exact manner.
 
\section{Conclusion} \label{sec:Con}
We derive analytical expressions for the MFVB's approximation error for a VAR with a conjugate prior. We also show that errors of a similar order hold for a VAR with independent priors based on a real data example. Our results confirm the claim that MFVB underestimates variances of the parameters and shows that the magnitude of underestimation (and also the KL divergence) change directly with the size of the model but inversely with the number of observations and the strength of the prior for variance. MFVB, at least for standard VARs, is as good as MCMC for estimating posterior means, and for forecasting \citep[our results and][]{koop2018variational}. MFVB also provides an ideal candidate for reciprocal importance sampling \citep[see] []{hajargasht2018accurate}. Estimation of variances can also be improved \citep[e.g.] []{giordano2017covariances}. 
\appendix
\section{Derivation of (3.5)} \label{3.5}
We start with the posterior kernel
\begin{align}
\begin{split}
  & \ln p(\mathbf{Y},\mathbf{\Gamma },{{\mathbf{\Sigma }}^{-1}})=C+\frac{1}{2}(T+p+\underline{\upsilon }-M-1)\ln \left| {{\mathbf{\Sigma }}^{-1}} \right| \\ 
 & -\frac{1}{2}tr\left\{ {{\mathbf{\Sigma }}^{-1}}[(\mathbf{Y}-\mathbf{X\Gamma }{)}'(\mathbf{Y}-\mathbf{X\Gamma }) \right.\ \left. \mathbf{+}(\mathbf{\Gamma }-\mathbf{\underline{\Gamma }}{)}'\,{{{\mathbf{\underline{V}}}}^{\mathbf{-}1}}(\mathbf{\Gamma }-\mathbf{\underline{\Gamma }})\mathbf{+\underline{S}}] \right\} 
 \end{split}
\end{align}
where C is a constant term. Taking expectation with respect to ${{\mathbf{\Sigma }}^{-1}}$ and using the completing of squares trick \citep[see e.g.][]{Karlsson2013} we can show:
\begin{align}
\begin{split}
  & q(\mathbf{\Gamma })\leftarrow \mathcal{MN}\{\mathbf{\bar{\Gamma }},{{\overline{{{\mathbf{\Sigma }}^{-1}}}}^{-1}},\mathbf{\bar{V}}\} \\ 
 & \quad \mathbf{\bar{V}=}{{[{{{\mathbf{\underline{V}}}}^{-1}}\mathbf{+{X}'X}]}^{-1}}\ \ \ \mathbf{\bar{\Gamma }=\bar{V}}[{{{\mathbf{\underline{V}}}}^{-1}}\mathbf{\underline{\Gamma }+{X}'Y}]
 \end{split}
\end{align}
Similarly, after taking expectations with respect to $\mathbf{\Gamma }$ we have
\begin{equation}
q({{\mathbf{\Sigma }}^{-1}})\leftarrow \mathcal{W}({{\mathbf{\bar{S}}}_{q}}^{-1},{{\bar{\upsilon }}_{q}})\quad \quad {{\bar{\upsilon }}_{q}}=T+p+\underline{\upsilon }                      \end{equation}      
\[{{\mathbf{\bar{S}}}_{q}}\mathbf{=}(\mathbf{Y}-\mathbf{X\bar{\Gamma }}{)}'(\mathbf{Y}-\mathbf{X\bar{\Gamma }})\mathbf{+\underline{S}+}(\mathbf{\bar{\Gamma }}-\mathbf{\underline{\Gamma }}{)}'{{\mathbf{\underline{V}}}^{-1}}(\mathbf{\bar{\Gamma }}-\mathbf{\underline{\Gamma }})+\mathbf{\Omega }\]   
where ${{\mathbf{\Omega }}_{ij}}=tr\left[ ({{{\mathbf{\underline{V}}}}^{-1}}\mathbf{+{X}'X})\operatorname{cov}({{\mathbf{\Gamma }}_{i}},{{\mathbf{\Gamma }}_{j}}) \right]$ and ${\mathbf{\Gamma }}_{i}$ 
and ${\mathbf{\Gamma }}_{j}$ are $i\text{-}th$ and $j\text{-}th$ columns of $\mathbf{\Gamma }$. To derive this we have used some results on expectations from Appendix-\ref{Dist}.\\                              
Normally, appropriate moments from (A.2) and (A.3) are iterated until convergence and then $q(\mathbf{\Gamma })q({{\mathbf{\Sigma }}^{-1}})$ is treated as the posterior but note from (A.2) that we can write 
\begin{equation*}
    \operatorname{cov}({{\mathbf{\Gamma }}_{i}},{{\mathbf{\Gamma }}_{j}}){{=}^{\ }}\bar{\sigma} _{ij}^{+}\mathbf{\bar{V}}
\end{equation*}
where $\bar{\sigma} _{ij}^{+}$s are elements of matrix ${{\overline{{{\mathbf{\Sigma }}^{-1}}}}^{\,-1}}$.\, Therefore
\begin{equation}
\mathbf{\Omega }={{\overline{{{\mathbf{\Sigma }}^{-1}}}}^{\,-1}}=\frac{p}{T+p+\underline{\upsilon }}{{\mathbf{\bar{S}}}_{q}}
\end{equation}
where the last equality comes from ${{\mathbf{\Sigma }}^{-1}}\sim{\ }\mathcal{W}({{\mathbf{\bar{S}}}_{q}}^{-1},{{\bar{\upsilon }}_{q}}).$ Substituting $\mathbf{\Omega }$ from (A.4) in (A.3) we obtain:  
\begin{align}
  & {{{\mathbf{\bar{S}}}}_{q}}\mathbf{=}(\mathbf{Y}-\mathbf{X\bar{\Gamma }}{)}'(\mathbf{Y}-\mathbf{X\bar{\Gamma }})\mathbf{+\underline{S}+}(\mathbf{\bar{\Gamma }}-\mathbf{\underline{\Gamma }}{)}'{{{\mathbf{\underline{V}}}}^{-1}}(\mathbf{\bar{\Gamma }}-\mathbf{\underline{\Gamma }})+\frac{p}{T+p+\underline{\upsilon }}{{{\mathbf{\bar{S}}}}_{q}} \\ 
 & \quad \ \ =\frac{{{{\bar{\upsilon }}}_{q}}}{{\bar{\upsilon }}}[(\mathbf{Y}-\mathbf{X\bar{\Gamma }}{)}'(\mathbf{Y}-\mathbf{X\bar{\Gamma }})\mathbf{+\underline{S}+}(\mathbf{\bar{\Gamma }}-\mathbf{\underline{\Gamma }}{)}'{{{\mathbf{\underline{V}}}}^{-1}}(\mathbf{\bar{\Gamma }}-\mathbf{\underline{\Gamma }})] \ 
 \end{align}
Note that starting with (A.2) then (A.6) and (A4), $q(\mathbf{\Gamma })$ and  $q({{\mathbf{\Sigma }}^{-1}})$ can be computed separately without a need for an iterative scheme.
\section{Derivation of $q(\mathbf{y}_{T+1}^{{}}|\mathbf{Y})$} \label{Der}
 
Start with ${{\mathbf{y}}_{T+1}}={{\left( \mathbf{x}_{T+1}^{{}}\mathbf{\Gamma } \right)}^{\prime }}+{{\mathbf{\varepsilon }}_{T+1}}$ then using (3.5) and (D.1) we have
\begin{align*}
\begin{split}
  & q(\mathbf{x}_{T+1}^{{}}\mathbf{\Gamma }|\mathbf{Y})=\mathcal{MN}\left( \mathbf{x}_{T+1}^{{}}\mathbf{\bar{\Gamma }},{{\overline{{{\mathbf{\Sigma }}^{-1}}}}^{-1}},\mathbf{x}_{T+1}^{{}}\mathbf{\bar{V}{x}'}_{T+1}^{{}} \right) \\ 
 & \quad \quad \quad \quad \quad =\mathcal{N}\left( \mathbf{x}_{T+1}^{{}}\mathbf{\bar{\Gamma }},{{\overline{{{\mathbf{\Sigma }}^{-1}}}}^{-1}}\otimes \mathbf{x}_{T+1}^{{}}\mathbf{\bar{V}{x}'}_{T+1}^{{}} \right) \\ 
 & \quad \quad \quad \quad \quad =\mathcal{N}\left( \mathbf{x}_{T+1}^{{}}\mathbf{\bar{\Gamma }},\frac{{\mathbf{\bar{S}}}}{{\bar{\upsilon }}}\mathbf{x}_{T+1}^{{}}\mathbf{\bar{V}{x}'}_{T+1}^{{}} \right) \
 \end{split}
\end{align*}
Using (3.5) and (D.4)
\[q({{\mathbf{\varepsilon }}_{T+1}}|\mathbf{Y})=\int{p({{\mathbf{\varepsilon }}_{T+1}}|{{\mathbf{\Sigma }}^{-1}},\mathbf{Y})}\,q({{\mathbf{\Sigma }}^{-1}}|\mathbf{Y})d{{\mathbf{\Sigma }}^{-1}}=\mathcal{T}(\mathbf{0},\frac{1}{{{{\bar{\upsilon }}}_{q}}}{{\mathbf{\bar{S}}}^{{}}}_{q},{{\bar{\upsilon }}_{q}})\]
using (D.4) again gives 
\[E({{\mathbf{\varepsilon }}_{T+1}}|\mathbf{Y})=\mathbf{0}\ \And \ Var({{\mathbf{\varepsilon }}_{T+1}}|\mathbf{Y})=\frac{{{{\mathbf{\bar{S}}}}_{q}}}{{{{\bar{\upsilon }}}_{q}}-2}=\frac{\bar{\upsilon }_{q}^{{}}}{({{{\bar{\upsilon }}}_{q}}-2)\bar{\upsilon }}\mathbf{\bar{S}}\]
Using (D.1) and (D.4) one more time:
\[E({{\mathbf{y}}_{T+1}}|\mathbf{Y})=E[(\mathbf{x}_{T+1}^{{}}\mathbf{\Gamma }{)}'|\mathbf{Y}]+E({{\mathbf{\varepsilon }}_{T+1}}|\mathbf{Y})=(\mathbf{x}_{T+1}^{{}}\mathbf{\bar{\Gamma }}{)}'\]
\begin{align*}
\begin{split}
  & Var({{\mathbf{y}}_{T+1}}|\mathbf{Y})=Var[(\mathbf{x}_{T+1}^{{}}\mathbf{\Gamma }{)}'|\mathbf{Y}]+Var({{\mathbf{\varepsilon }}_{T+1}}|\mathbf{Y}) \\ 
 & \quad \quad \quad \quad \ \ \ \ \ =(\mathbf{x}_{T+1}^{{}}\mathbf{\bar{V}{x}'}_{T+1}^{{}})\frac{{\mathbf{\bar{S}}}}{{\bar{\upsilon }}}+\frac{{{{\bar{\upsilon }}}_{q}}\mathbf{\bar{S}}}{({{{\bar{\upsilon }}}_{q}}-2)\bar{\upsilon }} \\ 
 & \quad \quad \quad \quad \ \ \ \ =\left( \mathbf{x}_{T+1}^{{}}\mathbf{\bar{V}{x}'}_{T+1}^{{}}+\frac{{{{\bar{\upsilon }}}_{q}}}{{{{\bar{\upsilon }}}_{q}}-2} \right)\frac{{\mathbf{\bar{S}}}}{{\bar{\upsilon }}} \
 \end{split}
\end{align*}

\section{Derivation of Posterior Modes} \label{Mode}
\textit{(i)}	Conjugate VAR:
Note that log-posterior can be written as\begin{align*}
  & \ln p(\mathbf{\Gamma },{{\mathbf{\Sigma }}^{-1}}|\mathbf{Y})=C+\frac{1}{2}(T+p+\underline{\upsilon }-M-1)\ln \left| {{\mathbf{\Sigma }}^{-1}} \right| \\ 
 & \quad \quad \quad \quad \quad -\frac{1}{2}tr\left\{ {{\mathbf{\Sigma }}^{-1}}[(\mathbf{\Gamma }-\mathbf{\bar{\Gamma }}{)}'{{{\mathbf{\underline{V}}}}^{-1}}(\mathbf{\Gamma }-\mathbf{\bar{\Gamma }}) \right.\ \left. +\mathbf{\bar{S}}] \right\} \\ 
\end{align*}
Maximize this with respect to $(\mathbf{\Gamma },{{\mathbf{\Sigma }}^{-1}})$
  \begin{align*}
      \frac{\partial \ln p}{\partial \mathbf{\Gamma }}=-{{\mathbf{\hat{\Sigma }}}^{-1}}{{\mathbf{\underline{V}}}^{-1}}(\mathbf{\hat{\Gamma }}-\mathbf{\bar{\Gamma }})=\mathbf{0}\Rightarrow \mathbf{\hat{\Gamma }}=\mathbf{\bar{\Gamma }}
     \end{align*}
      \begin{align*}
  \begin{split}
  & \frac{\partial \ln p}{\partial {{\mathbf{\Sigma }}^{-1}}}=\frac{1}{2}(T+p+\underline{\upsilon }-M-1)\mathbf{\hat{\Sigma }}-\frac{1}{2}[(\mathbf{\hat{\Gamma }}-\mathbf{\bar{\Gamma }}{)}'{{{\mathbf{\underline{V}}}}^{-1}}(\mathbf{\hat{\Gamma }}-\mathbf{\bar{\Gamma }})+\mathbf{\bar{S}}]=\mathbf{0} \\ 
 & \quad \quad \quad \Rightarrow \frac{1}{2}(T+p+\underline{\upsilon }-M-1)\mathbf{\hat{\Sigma }}-\frac{1}{2}\mathbf{\bar{S}}=\mathbf{0} \\ 
 & \quad \quad \quad \Rightarrow {{{\mathbf{\hat{\Sigma }}}}^{-1}}=(T+p+\underline{\upsilon }-M-1){{{\mathbf{\bar{S}}}}^{-1}}\quad \quad \quad \quad \quad \quad \quad \quad  \
 \end{split}
\end{align*}
where ($\mathbf{\hat{\Gamma }}$, ${{\mathbf{\hat{\Sigma }}}^{-1}}$) denotes the mode of posterior. To derive these results we have used some matrix derivative formulas from the Appendix-\ref{Dist}.   \\
\textit{(ii)}	Independent Prior VAR\\
Log-posterior:
\begin{align*}
\begin{split}
  & \ln p=C+\frac{1}{2}(T+\underline{\upsilon }-m-1)\ln \left| {{\mathbf{\Sigma }}^{-1}} \right|-\frac{1}{2}tr\left\{ {{{\mathbf{\underline{V}}}}^{-1}}(\mathbf{\beta }-\mathbf{\underline{\beta }})(\mathbf{\beta }-\mathbf{\underline{\beta }}{)}' \right\} \\ 
 & \quad \quad \quad -\frac{1}{2}tr\left\{ {{\mathbf{\Sigma }}^{-1}}[\mathbf{\underline{S}}+\sum\limits_{t=1}^{T}{({{\mathbf{y}}_{t}}-{{\mathbf{Z}}_{t}}\mathbf{\beta })({{\mathbf{y}}_{t}}-{{\mathbf{Z}}_{t}}\mathbf{\beta }{)}'}] \right\} \
 \end{split}
\end{align*}
Maximize this with respect to $(\mathbf{\Gamma },{{\mathbf{\Sigma }}^{-1}})$\\
\begin{align*}
\begin{split}
  & \frac{\partial \ln p}{\partial \mathbf{\beta }}=\sum\limits_{t=1}^{T}{[{{\mathbf{Z}}^{\prime }}{{{\mathbf{\hat{\Sigma }}}}^{-1}}{{\mathbf{y}}_{t}}-{{\mathbf{Z}}_{t}}^{\prime }{{{\mathbf{\hat{\Sigma }}}}^{-1}}{{\mathbf{Z}}_{t}}\mathbf{\beta })]}-{{{\mathbf{\underline{V}}}}^{-1}}(\mathbf{\hat{\beta }}-\mathbf{\underline{\beta }})=\mathbf{0} \\ 
 & \quad \Rightarrow \mathbf{\hat{\beta }}=[{{{\mathbf{\underline{V}}}}^{-1}}+\sum\limits_{t=1}^{T}{\mathbf{Z}{{_{t}^{{}}}^{\prime }}}{{{\mathbf{\hat{\Sigma }}}}^{-1}}{{\mathbf{Z}}_{t}}]^{-1}[{{{\mathbf{\underline{V}}}}^{-1}}\mathbf{\underline{\beta }}+\sum\limits_{t=1}^{T}{\mathbf{Z}{{_{t}^{{}}}^{\prime }}}{{{\mathbf{\hat{\Sigma }}}}^{-1}}{{\mathbf{y}}_{t}}]\ \ \quad \quad \quad \text{(A.7)} \\
 \end{split}
\end{align*}
\begin{align*}
\begin{split}
  & \frac{\partial \ln p}{\partial {{\mathbf{\Sigma }}^{-1}}}=\frac{1}{2}(T+\underline{\upsilon }-M-1)\mathbf{\hat{\Sigma }}-\frac{1}{2}[\mathbf{\underline{S}}+\sum\limits_{t=1}^{T}{({{\mathbf{y}}_{t}}-{{\mathbf{Z}}_{t}}\mathbf{\hat{\beta }})({{\mathbf{y}}_{t}}-{{\mathbf{Z}}_{t}}\mathbf{\hat{\beta }}{)}'}]=\mathbf{0} \\ 
 & \quad \quad \Rightarrow \frac{1}{2}(T+\underline{\upsilon }-M-1)\mathbf{\hat{\Sigma }}-\frac{1}{2}[\mathbf{\underline{S}}+\sum\limits_{t=1}^{T}{({{\mathbf{y}}_{t}}-{{\mathbf{Z}}_{t}}\mathbf{\hat{\beta }})({{\mathbf{y}}_{t}}-{{\mathbf{Z}}_{t}}\mathbf{\hat{\beta }}{)}'}]=\mathbf{0} \\ 
 & \quad \quad \Rightarrow {{{\mathbf{\hat{\Sigma }}}}^{-1}}=(T+\underline{\upsilon }-M-1)[\mathbf{\underline{S}}+\sum\limits_{t=1}^{T}{({{\mathbf{y}}_{t}}-{{\mathbf{Z}}_{t}}\mathbf{\hat{\beta }})({{\mathbf{y}}_{t}}-{{\mathbf{Z}}_{t}}\mathbf{\hat{\beta }}{)}'}]^{-1}\quad \text{(A.8)} 
 \end{split}
\end{align*}
To derive these results, we have used some matrix derivative formulas from Appendix-\ref{Dist}. The mode of the exact posterior i.e. [$\mathbf{\hat{\beta }}$,${{\mathbf{\hat{\Sigma }}}^{-1}}$] can be obtained by iterating over equations (A.7) and (A.8). 
 In contrast, the mode of the MFVB posterior can be obtained (using 3.9 and known results on modes of normal and Wishart distributions) by iterating over the following equations
\begin{align*}
\begin{split}
  & \mathbf{\bar{V}}_q=[{{{\mathbf{\underline{V}}}}^{-1}}+\sum\limits_{t=1}^{T}{\mathbf{Z}{{_{t}^{{}}}^{\prime }}}\lambda \mathbf{\hat{\Sigma }}_{q}^{-1}{{\mathbf{Z}}_{t}}]^{-1},\quad {{{\mathbf{\hat{\beta }}}}_{q}}={{{\mathbf{\bar{V}}}}_{q}}[{{{\mathbf{\underline{V}}}}^{-1}}\mathbf{\underline{\beta }}+\sum\limits_{t=1}^{T}{\mathbf{Z}{{_{t}^{{}}}^{\prime }}}\lambda \mathbf{\hat{\Sigma }}_{q}^{-1}{{\mathbf{y}}_{t}}]\  \\ 
 & \mathbf{\hat{\Sigma }}_{q}^{-1}=(T+\underline{\upsilon }-M-1)[\mathbf{\underline{S}}+\sum\limits_{t=1}^{T}{({{\mathbf{y}}_{t}}-{{\mathbf{Z}}_{t}}{{{\mathbf{\hat{\beta }}}}_{q}})({{\mathbf{y}}_{t}}-{{\mathbf{Z}}_{t}}{{{\mathbf{\hat{\beta }}}}_{q}}{)}'}+{{\mathbf{Z}}_{t}}{{{\mathbf{\bar{V}}}}_{q}}{{{\mathbf{{Z}'}}}_{t}}]^{-1} 
\end{split}
\end{align*}
where  $\lambda =1+\dfrac{M+1}{T+\underline{\upsilon }-M-1}$ .

\section{Distributions, Matrix Derivatives and Expectations} \label{Dist}
\textit{\textbf{Distributions:}}\\
Matricvariate Normal Distribution: if ${{\mathbf{X}}_{pm}}\sim{\ }\mathcal{MN}\{\mathbf{\bar{X}},\mathbf{\Sigma },\mathbf{V}\}$ then
  \begin{align}
  \begin{split}
  & \text{vec}(\mathbf{X})\sim{\ }\mathcal{N}\{\text{vec}(\mathbf{\bar{X}}),\mathbf{\Sigma }\otimes \mathbf{V}\} \\ 
 & E(\mathbf{X})=\mathbf{\bar{X}}\quad \text{VAR}\{vec(\mathbf{X})\}=\mathbf{\Sigma }\otimes \mathbf{V} \\ 
 & {{\mathbf{A}}_{qp}}{{\mathbf{X}}_{pm}}\sim{\ }\mathcal{MN}\{\mathbf{A\bar{X}},\mathbf{\Sigma },\mathbf{AV{A}'}\}
 \end{split}
\end{align}
Wishart Distribution: if $\mathbf{\Sigma }\sim{\ }\mathcal{W}\{\mathbf{S},\upsilon \}$ then
\begin{equation}
 E(\mathbf{\Sigma })=\upsilon \mathbf{S}\ \ \And \text{VAR}\{{{\sigma }_{ij}}\}=\upsilon \left( s{{_{ij}^{{}}}^{2}}+s_{ii}^{{}}s_{jj}^{{}} \right)
\end{equation}  
Matricvariate t-distribution:
if ${{\mathbf{X}}_{pq}}\sim{\ }\mathcal{MT}\{\mathbf{\bar{X}},{{\mathbf{S}}_{q}},{{\mathbf{V}}_{p}},\upsilon \}$ then
 \begin{align}
 \begin{split}
  & E(\mathbf{X})=\mathbf{\bar{X}}\quad \text{VAR}\{vec(\mathbf{X})\}=\frac{\mathbf{S}\otimes \mathbf{V}}{\upsilon -q-1} \\ 
 & \text{ }\mathbf{X}|\mathbf{\Sigma }\sim{\ }\mathcal{MN}\{\mathbf{\bar{X}},\mathbf{\Sigma },\mathbf{V}\}\And \mathbf{\Sigma }\sim{\ }W\{\mathbf{S},\upsilon \} \\ 
 & \quad \quad \ \ \Rightarrow {{\mathbf{X}}_{pq}}\sim{\ }\mathcal{MT}\{\mathbf{\bar{X}},\mathbf{S},\mathbf{V},\upsilon \}  
 \end{split}
 \end{align}
Multivariate t-distribution: if ${{\mathbf{x}}_{q\times 1}}\sim{\ }\mathcal{T}\{\mathbf{x},\mathbf{S},\upsilon \}$ then 
\begin{align}
\begin{split}
  & E(\mathbf{x})=\mathbf{\bar{x}}\quad \text{VAR}\{\mathbf{x}\}=\frac{\upsilon \mathbf{S}}{\upsilon -2}\quad \quad  \\ 
 & \mathbf{x}|\mathbf{\Sigma }\sim{\ }\mathcal{N}\{\mathbf{\bar{x}},\mathbf{\Sigma },\mathbf{V}\}\And \mathbf{\Sigma }\sim{\ }\mathcal{W}\{\mathbf{S},\upsilon \} \\ 
 & \quad \quad \Rightarrow \mathbf{x}\sim{\ }\mathcal{T}\{\mathbf{\bar{x}},\frac{1}{\upsilon }\mathbf{\bar{S}},\upsilon \} 
 \end{split}
\end{align}
\textit{\textbf{Some Matrix Derivatives}}: 
\begin{equation}
  \frac{\partial \ln |{{\mathbf{X}}_{m\times m}}|}{\partial \mathbf{X}}={{\mathbf{{X}'}}^{-1}}
\end{equation}
\begin{equation}
   \frac{\partial tr\{{{\mathbf{A}}_{m\times m}}{{\mathbf{X}}_{m\times m}}\}}{\partial \mathbf{X}}=\mathbf{{A}'} 
\end{equation}
\begin{equation}
 \frac{\partial tr\{\mathbf{A_{m\times m}X_{m\times m}'B_{m\times m}X_{m\times m}}\}}{\partial \mathbf{X_{m\times m}}}=\mathbf{ABX}+\mathbf{{A}'{B}'X} \\ 
\end{equation} 
\begin{equation}
\frac{\partial tr\{\mathbf{AX{X}'}\}}{\partial \mathbf{X}}=\mathbf{AX}+\mathbf{{A}'X}
\end{equation}
\begin{equation}
\frac{\partial tr\{{{\mathbf{A}}_{m\times n}}{{\mathbf{B}}_{n\times k}}{{\mathbf{X}}_{k\times l}}{{\mathbf{C}}_{l\times m}}\}}{\partial \mathbf{X}}={{\mathbf{{B}'}}_{k\times n}}{{\mathbf{{A}'}}_{n\times m}}{{\mathbf{C}}_{m\times l}}
\end{equation}
\begin{equation}
\frac{\partial tr\{{{\mathbf{A}}_{m\,m}}{{\mathbf{B}}_{m\,k}}{{\mathbf{x}}_{k\,1}}{{{\mathbf{{x}'}}}_{1\, k}}{{{\mathbf{{B}'}}}_{k\,m}}\}}{\partial \mathbf{x}}={{\mathbf{{B}'}}_{k\,m}}{{\mathbf{{A}'}}_{m\,m}}{{\mathbf{B}}_{m\,k}}{{\mathbf{x}}_{k\,1}}+{{\mathbf{{B}'}}_{k\,m}}{{\mathbf{A}}_{m\,m}}{{\mathbf{B}}_{m\,k}}{{\mathbf{x}}_{k\,1}}
\end{equation}
\textit{\textbf{Some Results on Expectations}}:\\
\hspace*{0.5cm} Let $E(\mathbf{x})=\mathbf{\bar{x}}$ and $Var(\mathbf{x})=\mathbf{\Sigma }$ then
\begin{equation}
E({{\mathbf{{x}'}}_{1\times q}}{{\mathbf{A}}_{q\times q}}{{\mathbf{x}}_{q\times 1}})=\mathbf{\bar{{x}'}A\bar{x}}+tr[\mathbf{A\Sigma }]
\end{equation}
\begin{equation}
E({{\mathbf{{x}'}}_{1\times q}}{{\mathbf{A}}_{q\times m}}{{\mathbf{y}}_{m\times 1}})=\mathbf{\bar{{x}'}A\bar{y}}+tr[\mathbf{A}Cov(\mathbf{x},\mathbf{y})]
\end{equation}
\begin{equation}
E(\mathbf{x{x}'})=\mathbf{\bar{x}}\,\mathbf{{\bar{x}}'}-\mathbf{\Sigma }
\end{equation}
\begin{equation}
E({{\mathbf{{X}'}}_{q\times m}}{{\mathbf{A}}_{m\times m}}{{\mathbf{X}}_{m\times q}})={{\mathbf{\bar{{X}'}}}_{{}}}\mathbf{A\bar{X}}+\mathbf{\Omega }\ 
\end{equation}
where ${{\mathbf{\Omega }}_{ij}}=tr\{Cov({{\mathbf{x}}_{i}},{{\mathbf{x}}_{j}})\mathbf{A}\}$, ${{\mathbf{x}}_{i}}$ and ${{\mathbf{x}}_{j}}$ are i-th and    j-th columns of $\mathbf{X}$ for $i,j=1,...,q$.

\bibliographystyle{ba}
\bibliography{sample}


\end{document}